\documentclass[letterpaper, 10 pt,
conference]{ieeeconf} 

\IEEEoverridecommandlockouts 

\usepackage{amsmath} 
\usepackage{amssymb} 
\usepackage{xcolor}
\usepackage{algpseudocode}
\usepackage{algorithmicx}
\usepackage{algorithm}
\usepackage{lmodern}
\usepackage{siunitx}

\usepackage{tikz}
\usepackage{enumerate}

\usepackage{amsthm}

\theoremstyle{definition}

\newcounter{defcount}
\renewcommand*\thedefcount{\textbf{\arabic{defcount}}}
\stepcounter{defcount}

\newcommand{\myDef}[0]{\\[-0.15 mm]\noindent\textbf{Def \thedefcount.~}\addtocounter{defcount}{1}}

\def\figref#1{Fig.~\ref{#1}}

\def\eqref#1{Eq.~(\ref{#1})}

\newcommand\etal{et al.}

\title{\LARGE \bf A Constraint Programming Approach to Simultaneous Task Allocation and Motion Scheduling for Industrial Dual-Arm Manipulation Tasks}

\author{
Jan Kristof Behrens$^{1}$\thanks{$^{1}$Robert Bosch GmbH, Corporate Research, Renningen, Germany, \texttt{behrens.jk@gmail.com}, \texttt{ralph.lange@de.bosch.com}}
\and 
Ralph Lange$^{1}$ 
\and
Masoumeh Mansouri$^{2}$\thanks{$^{2}$\"{O}rebro University, Sweden, \texttt{masoumeh.mansouri@oru.se}}
}

\begin{document}
\maketitle
\thispagestyle{empty} \pagestyle{empty}


\begin{abstract} 
Modern lightweight dual-arm robots bring the physical capabilities to quickly take over tasks at typical industrial workplaces designed for workers. In times of mass-customization, low setup times including the instructing/specifying of new tasks are crucial to stay competitive. We propose a constraint programming approach to simultaneous task allocation and motion scheduling for such industrial manipulation and assembly tasks. The proposed approach covers dual-arm and even multi-arm robots as well as connected machines. The key concept are Ordered Visiting Constraints, a descriptive and extensible model to specify such tasks with their spatiotemporal requirements and task-specific combinatorial or ordering constraints. Our solver integrates such task models and robot motion models into constraint optimization problems and solves them efficiently using various heuristics to produce makespan-optimized robot programs. The proposed task  
model is robot independent and thus can easily be deployed to other robotic platforms. Flexibility and portability of our proposed model is validated through several experiments on different simulated robot platforms. We benchmarked our search strategy against a general-purpose heuristic. 
For large manipulation tasks with 200 objects, our solver implemented using Google's Operations Research tools and ROS requires less than a minute to compute usable plans.

\end{abstract}
\vspace{-0.2cm}

\section{Introduction}





\noindent Modern lightweight dual-arm robots such as the ABB YuMi or the KaWaDa Nextage are engineered in the style of a human torso to be easily applicable in industrial workplaces designed for workers. These types of robots are an answer to the demand for flexible, cost-efficient production of customer-driven product variants and small lot sizes.

Such flexible production requires fast methods to specify new tasks for these robots. Classical teach-in by means of fixed poses and paths is not appropriate. With the capabilities of today's perception systems, which can detect and localize workpieces, boxes, and tools automatically in typical workplaces, and a formalized goal or high-level task specification, the manual teach-in may be replaced by automated planning -- in principle. Optimal planning involves three aspects: (a) \emph{task planning} of the necessary steps and actions to achieve the overall goal/task, (b) \emph{scheduling} of these steps and actions, and (c) \emph{motion planning} for each step and action.
\begin{figure}[!t]
\centering
\includegraphics[width=0.65\linewidth]{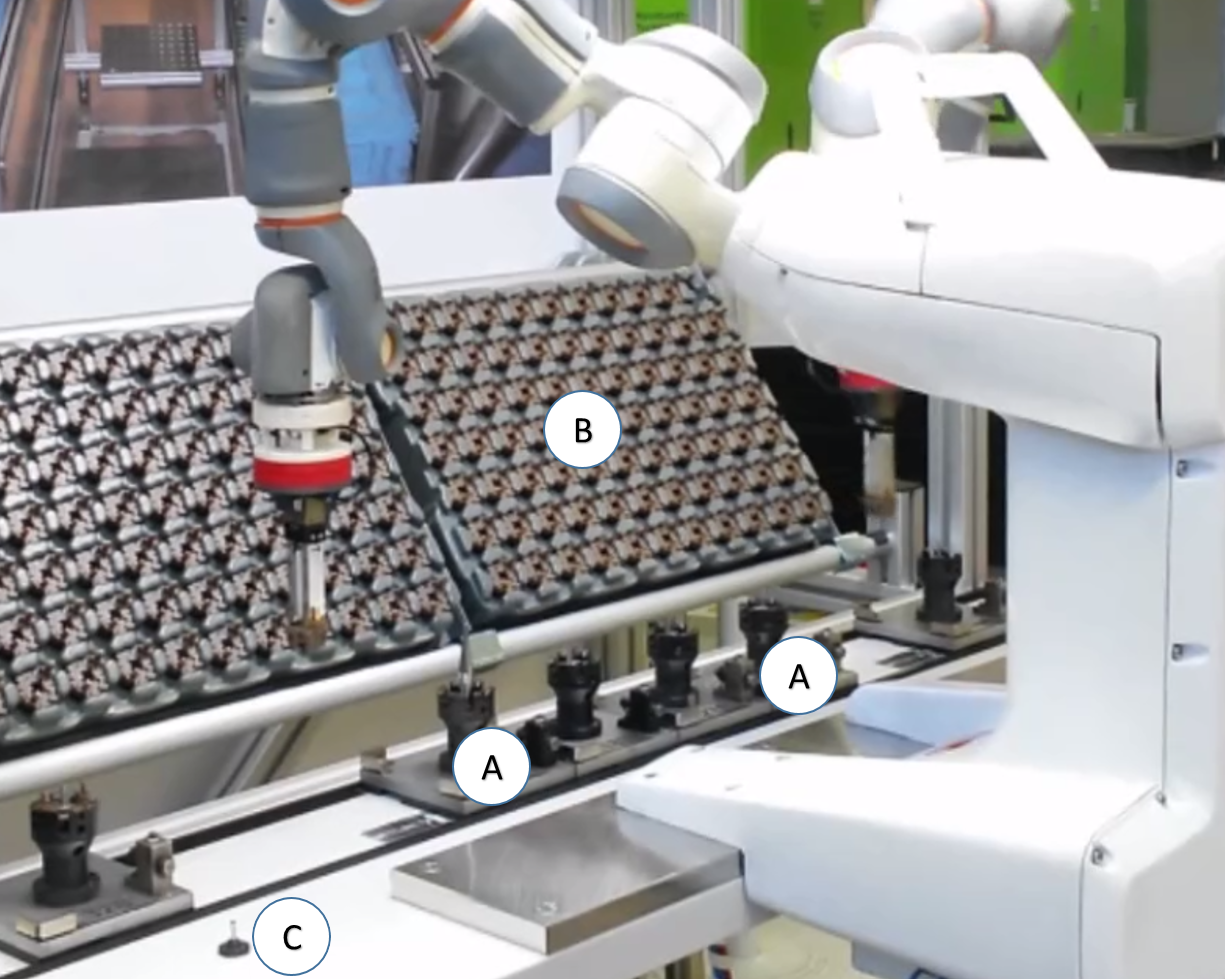}
\vspace{-0.2cm}
\caption{Assembling of wiper motors with a dual-arm robot. The robot picks a tool from (C), places it on the shaft of the rotor of an electric motor in the workpiece holder (A), picks an electric interface, supplied in a container (B) and places it on (A).}
\vspace{-0.85cm}
\label{fig:use-case}
\end{figure}
Dual-arm manipulation further requires to decide about (d) the \emph{allocation} of task steps and actions to the individual arms. Moreover, the complexity of scheduling and motion planning is increased heavily, due to the necessity to closely coordinate the manipulators to prevent self-collisions of the robot.

All four aspects -- task planning, scheduling, allocation and motion planning -- are closely interrelated. Ideally, to achieve optimal plans with regard to the makespan (production time) or similar objectives, they have to be considered in one coherent formalism and planning algorithm. In the last years, significant progress has been made to closely couple task planning with motion planning by passing feedback from motion planning to task planning (e.g., \cite{garrett_ffrob:_2018, cambon_hybrid_2009,lagriffoul_combining_2016,lozano-perez_constraint-based_2014,dantam_incremental_2018}), but research is still far from an ideal solution.

In many industrial use-cases, task planning is not required as the necessary steps and actions to process and assemble a workpiece are already given in digital form. That is, we already have an abstract plan, but with a number of unknowns and degrees of freedom in terms of the three aspects scheduling, allocation and motion planning. Computing an optimal, executable plan requires to treat these aspects in a highly integrated and coherent manner, which we refer to as \emph{simultaneous task allocation and motion scheduling} (STAAMS). An optimal plan depends not only on the motion of the manipulators but also on the order in which a workpiece is assembled, the order in which the components are taken from boxes or conveyor belts, in which they are processed by other machines, etc. -- in particular, if connected systems or machines impose temporal constraints. The number of actions to be scheduled can be very high which results in big combinatorial complexity. Moreover, a suitable STAAMS solver has to consider different assignments of subtasks to arms, while taking the individual working ranges into account as well as task steps in which the arms have to cooperate.

In this paper, we propose a flexible model and solver for STAAMS for multi-arm robots in industrial use-cases. The proposed model and solver are based on constraint programming (CP) and constraint optimization, respectively. In detail, \textbf{our contributions} are as follows:

\vspace{0.05cm}
	 \noindent\textbf{1)} For specifying the abstract task decomposition of a STAAMS problem, we propose a novel and intuitive model primitive named \emph{Ordered Visiting Constraints} (OVC). The OVC concept is developed out of the observation that many production steps can be described concisely by sequences of actions (e.g. drilling, picking, welding or joining) to be performed with one of the robot arms at given locations, with ordering or temporal constraints between them.
	
	\noindent \textbf{2)} For the robot motions, we propose a model of \emph{time-scalable motion series} that can be directly integrated with constraint-based scheduling, utilizing the fact that many industrial workplaces provide a controlled and unobstructed environment in which motion planning can be performed using precomputed roadmaps.
	
	\noindent \textbf{3)} We propose an advanced CP concept named \emph{Connection Variables} to link the two submodels -- the OVC-based task model and the motion model -- into an unified STAAMS problem model. Connection Variables are a special kind of CP meta variables on the indicies of other CP variables. At the same time, we explain how this modularity allows to port a given OVC-based task model to different workplace layouts and robots.
	
	\noindent \textbf{4)} We present an adaptable solver, which allows for fine-grained user control over the different constraint optimization techniques to compute an almost-optimal plan for typical STAAMS problem sizes in few seconds.
  
The remainder of this paper is organized as follows: We present an analysis of typical industrial use-cases in Sec.~\ref{sec:use-case analysis} before we discuss related work in Sec.~\ref{sec:related-work}. Our main contributions, the STAAMS model with the OVC-based task model and the motion model as well as the corresponding solver, are presented in the Sections~\ref{sec:formalism} and \ref{sec:ovc-planning}, respectively. We show the scalability and portability of the proposed system and compare it to pure time-scaling in Sec.~\ref{sec:evaluation}. The paper is concluded in Sec.~\ref{sec:conclusion}.



\section{Use-case Analysis and Problem Definition}\label{sec:use-case analysis}
\noindent In this section, we describe two typical industrial use-cases for dual-arm robots, followed by an analysis of characteristic properties and prevalent concepts. These properties and the concept of Ordered Visiting Actions serves as basis for the design of our STAAMS model in Section~\ref{sec:formalism}.

\subsection{Use-Cases}\label{subsec:use-case analysis:use-case}


\textbf{Sorting objects.} The robot has to pick up colored objects from the table and place them depending on their color in one of two containers. All parts on the table are reachable by both arms. The containers are only reachable by either of the arms (see Fig.~\ref{fig:t-vs-q-a}) so that an object's color defines the arm that has to pick this object. This use-case inspired by Kimmel \etal~\cite{kimmel_scheduling_2016} will serve as a reference use-case for the evaluation.

\textbf{Injection molding.} Parts have to be taken from a source container and inserted into an injection molding machine. When the molding process is finished, the parts have to be taken from the machine and placed under a camera for visual inspection and hold into a fixture for an electrical check. The latter requires to press a button simultaneously to start the check. While the molding machine may process two parts simultaneously, the visual and electrical checks can only process one part at a time. Finally, the finished parts are placed in another container. Full containers have to be placed for collection.

\subsection{Analysis}\label{subsec:analysis}
\noindent These industrial use-cases show several characteristic properties:

\textbf{Controlled environment.} Industrial workplaces provide by design a controlled and unobstructed environment. Therefore, we assume that all object locations and possible placements are known in advance, which allows for offline pre-calculation of motion roadmaps and collision tables. Furthermore, we may assume the absence of external interferences such as humans.

\textbf{Unobstructed workspace.} We assume that relevant objects never obstruct each other. This implies that there exists a collision-free subset of the workspace that does not alter over time and allows to reach all relevant object locations with at least one robot arm. For example this applies to drilling, riveting, welding, glueing, and assembling of small parts. As a consequence, we do not require a complex scene graph (cf.~\cite{blumenthal_scene_2013}) that tracks geometric relations between all objects in the workspace.

\textbf{Ordered Visiting Actions.} Suitable plans for these use-cases may be specified as a series of motions (per arm) to visit relevant locations in the workspace. At each location, the manipulator may perform local actions such as screw in a screw or picking an object from a container, which -- for our scheduling purposes -- can be abstracted as constraint on the visiting duration at that location. While the overall order of actions may be changed, some actions like pick-and-place are subject to a partial ordering and are therefore considered as an entity. We refer to such entities as \emph{Ordered Visiting Actions} (OVA) in the following. OVAs may be used to model many advanced tasks such as joining, welding, sorting, inspecting, drilling, and milling.

\textbf{Temporal dependencies.} Often, there are additional temporal dependencies between OVAs. For example, molding has to precede the visual and electrical checks in the molding use-case and in the motor assembly use-case (as shown in Fig.~\ref{fig:use-case}) the temporal dependencies are given by the assembly sequence for each motor. 
In the sorting use-case, there are no temporal dependencies between the OVAs per se. Yet, each arm can transport only a single object at a time, i.e. the gripper is a reservable resource, which requires to schedule the OVAs per arm.


\textbf{Active components.} Another important observation is that processing stations in the workspace may also take on different configurations, just as the robot arms. An example is the door of the molding machine in the second use-case. We refer to such stations and all active robot components together as \emph{active components}.

\subsection{STAAMS Problem Definition}\label{subsec:STAAMS-definition}
\noindent Based on this analysis, STAAMS can be formulated as the anonymous variant of multi-agent path finding problem, combined with target assignments (TAPF) given by partially-ordered sets of subtasks/actions with spatiotemporal, combinatorial and ordering constraints on predefined locations (6DoF poses); and is NP-hard~\cite{sevn-ijcai-ws-16}.


\section{Related Work} \label{sec:related-work}
\noindent The state-of-the-art optimal TAPF method~\cite{Ma:2016:OTA:2936924.2937092} cannot solve STAAMS problems in general, as it does not compute kinematically feasible motions for agents, nor can it be applied in cases requiring ordering decisions about task assignments. Online methods for multi-agent task assignment and scheduling algorithms have been developed for small-sized teams of agents, and highly flexible against execution uncertainty~\cite{DBLP:conf/aips/ShahCW09}. Multi-robot task allocation with temporal/ordering constraints has been studied in the context of integrating auction-based methods with Simple Temporal Problems~\cite{NUNES201755}. These methods, however, do not account for conflicting spatial interactions, as needed, for example, in dual-arm manipulations. The applications of CP to multi-robot task planning and scheduling often use a simplified robot motion model, and ignore the cost of spatial interaction among robots in the scheduling process~\cite{DBLP:conf/cp/BoothNB16}. 



The motion planning and scheduling sub-problems of STAAMS can be seen as a multi-robot motion planning problem. State-of-the-art approaches tackling this problem often do not address the task/goal allocation problem, i.e., goals/tasks are assumed to be given. LaValle~\cite{lavalle_part_2006} formulates the motion scheduling problem in a joint configuration space. \emph{Prioritized planning} assigns an order to the robots (arms) according to which their movements are planned (e.g., \cite{kurosu_simultaneous_2017}). In \emph{fixed-path planning} -- also referred to as \emph{time-scaling} -- only timings are adjusted to prevent collisions \cite{odonnell_deadlock-free_1989}. In \emph{fixed-roadmap planning}, topological graphs are used to both plan the paths and adjust the timings. Our proposed method falls in the third category. A fixed-roadmap is particularly suitable for industrial settings, as the environments are not often subject to change.  Kimmel et al.~\cite{kimmel_scheduling_2016} employ a time-scaling approach to schedule two given sequences of pick-and-place tasks. Time-scaling problems can be modeled easily as a special-case with our STAAMS model. Our approach performs such time-scaling in its last step. In Sec.~\ref{sec:evaluation}, we compare our simultaneous task allocation and motion scheduling approach with pure time-scaling using an experimental setup in the style of the one used by Kimmel et al.



Alatartsev et al.~\cite{alatartsev_robotic_2015} present a survey about the task sequencing problem for industrial robots, where sources for execution variants are systematically identified for a given task specification (e.g., multiple inverse kinematic solutions, partial ordering) and optimized based on various cost functions. The survey, however, lacks the coverage for tasks that are applicable for multi-arm robots. Kolakowska et al. \cite{kolakowska_constraint_2014} schedule paint strokes by ignoring the dependency between the ordering of the strokes and their motions.  This approach is not generalizable to multi-robot scenarios, as this dependency cannot be ignored due to robot-robot collisions. Representing task orders can be done via hierarchical task networks (HTN)~\cite{Ghallab:2016:APA:3073924}. However, HTN would not by itself be capable of generating the orderings in the plan in a way that is optimized, or even feasible from the point of view of the robot's geometry or motions.


In our approach, we employ CP to model the abstract task specification and the robot motion. Similarly, Ejenstam et al.~\cite{ejenstam_implementing_2014} use CP to solve the problem of dual-arm manipulation planning and cell layout optimization via a coarse discretization of the workspace. Conversely, we create dense roadmaps to enable the close coordination of arms, thus allow more parallel movements of arms. Kurosu et al.~\cite{kurosu_simultaneous_2017} describe a decoupled MILP-based approach to solve a STAAMS, where the motion planner is prone to fail due to simplified motion and cost models used in the one-shot MILP formulation. This is not the case for us, as a single CP solver finds a mutually feasible solution for all sub-problems.

In our previous work \cite{behrens2017icra}, we hand-coded the full requirements of the robot and its workspace in the MiniZinc language. In this paper, we introduce a coherent formalism which allows to model the robot and workspace as well as an abstract task plan and its invariants. We propose OVCs as task model primitives and time-scalable motion series as motion model primitives. Also, we provide automated procedures to create the data objects such as the roadmaps for motion planning, which may be created automatically from 3D sensor data and cover the full workspace of a real robot. Most important, our planning system uses carefully chosen and evaluated variable ordering and value selection heuristics for efficient planning (see Fig.~\ref{fig:strategy_statistics}). For the implementation, we used the Google Operation Research tools~\cite{google_2018}, which (in contrast to MiniZinc solvers) allow fine-grained definition of the search strategy.

\section{Modeling STAAMS Problems with OVCs}\label{sec:formalism}

\begin{figure}[tpb]
\centering
\includegraphics[width=0.9\linewidth]{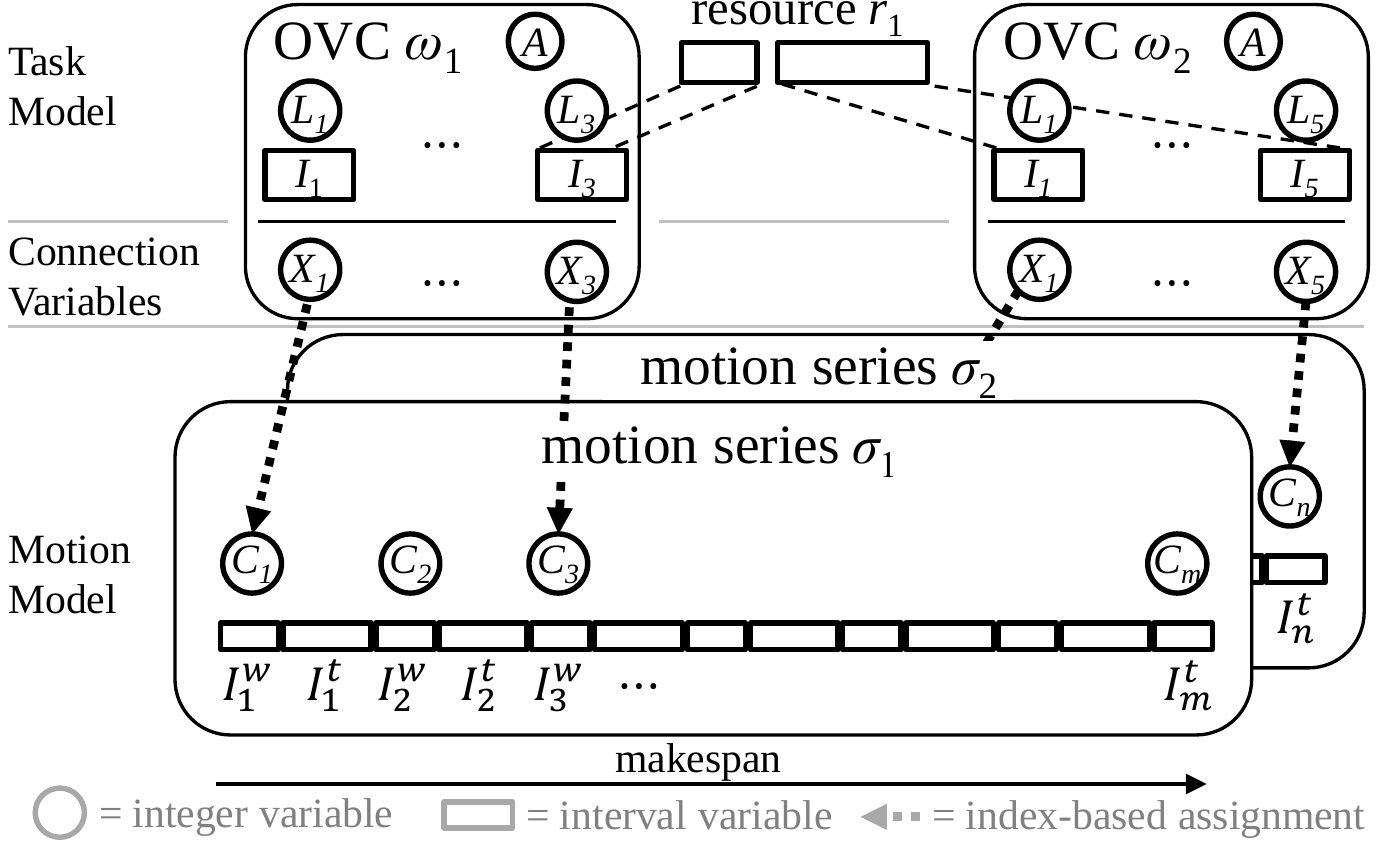}
\vspace{-0.4cm}
\caption{Overview of our CP-based STAAMS model}
\vspace{-0.6cm}
\label{fig:Overview to our model}
\end{figure}

\noindent In this section, we present our formalism for specifying STAAMS problems as constraint programs using OVCs. Our model consists of two submodels named \emph{task model} and \emph{motion model}. As illustrated in Figure~\ref{fig:Overview to our model}, the task model is independent of any kinematic details and actual trajectories. Conversely, the motion model represents the trajectories of all active components independent of any task information. Both models are linked through \emph{Connection Variables}, a special kind of CP variable explained below.

Next, we explain both submodels and then the Connection Variable mechanism. For readability, we write constants or values as lowercase Latin letters and constraint variables as capital letters. Compounds of constraint variables are denoted with small Greek letters.

\subsection{Task Model}
\noindent The first and most important element of the task model are OVCs, which can be considered as variable, constraint-based blueprints of OVAs. An OVC consists of four sets of CP variables modeling \emph{primitive actions} (e.g. pick, place, drill, etc.) to be executed at certain \emph{locations} in the workplace within certain \emph{time intervals} by an \emph{active component}.  The locations $\mathbb{L}$ are a finite set of 6DoF poses of interest in the workplace -- in particular possible object placements in containers and workpiece holders -- in a common reference system.
\myDef{Formally, an \textbf{OVC}}\label{def:ovc} is a tuple
\vspace{-0.25cm}
\begin{align*}
\omega &= (A, [P_1, ..., P_l], [L_1, ..., L_l],[I_1, ..., I_l], C_{\textrm{intra}})\textrm{.}
\end{align*}

\vspace{-0.25cm}
\noindent The variables $P_j$ represent the primitive actions, the variables $L_j \in \mathbb{L}$ describe the locations, and the variables $I_j$ model the time intervals. The variable $A$ represents the active component to be used. A triple $P_j$, $L_j$ and $I_j$ denotes that active component $A$ shall perform action $P_j$ during time $I_j$ at location $L_j$.

In $C_{\textrm{intra}}$, arbitrary constraints on and between these variables can be specified. In particular, each $P_j$ can be constrained to a specific primitive action to be executed. Similarly, each $L_j$ is typically constrained to one or few specific locations or specific location combinations for all location variables. Also, quantitative temporal constraints on the time intervals may be given.

The task model also allows for arbitrary constraints between OVCs, named \emph{inter-OVC constraints} $C_{\textrm{inter}}$. Typical examples are temporal constraints between OVCs (e.g., for synchronization or ordering of OVCs
or combinatorial constraints -- e.g., to distribute $m$ locations amongst $n$ OVCs). In the following, we refer to the set of all OVCs as $\Omega$.

The second element of the task model are \emph{resources} which describe abstract or physical objects such as tool, workpiece holders or robot grippers. A resource $r$ can be reserved exclusively for arbitrary time intervals. Typically, reservations are defined the by referencing start or end variables of interval variables of those OVCs that require this resource.

\begin{figure}
\centering
\includegraphics[width=0.85\linewidth]{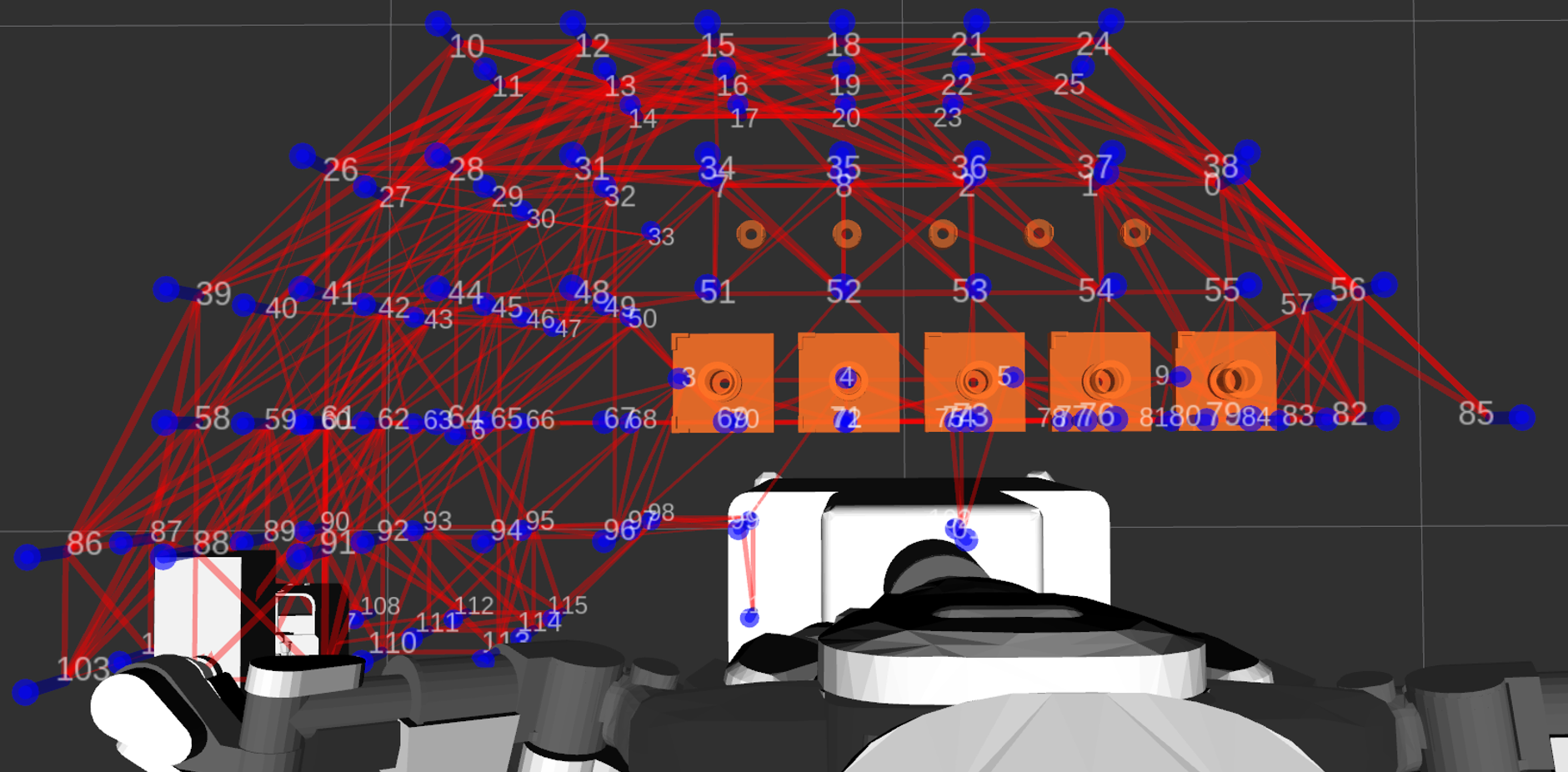}
\caption{A roadmap for the left arm of a KaWaDa Nextage robot.}
\vspace{-0.5cm}
\label{fig:roadmap_screenshot}
\end{figure}

\subsection{Motion Model}
\noindent The motion model represents the trajectories of all active components -- independent of any task information -- as \emph{motion series} consisting of configuration variables -- in the respective joint space of the active component -- with time interval variables for the transition in-between. To be able to model the motion with CP, a roadmap-based approach is used (cf.~for example \cite{kavraki_probabilistic_1996}). The roadmap of an active component is a sufficiently dense sampling of the joint space, where the nodes are joint configurations and the edges represent short collision-free motions between them (cf. Fig.~\ref{fig:roadmap_screenshot}). In particular, the roadmap contains one or more nodes for each location $l \in \mathbb{L}$ in reach of the active component.
\myDef{Hence, the \textbf{motion series}}\label{def:motion-series} $\sigma_a$ of an active component $a$ is a sequence of $m$ configuration variables and a sequence of $2m-1$ interval variables
\vspace{-0.25cm}
\begin{align*}
\sigma_a = ([C_1, \ldots, C_m], [I^\textrm{w}_1, I^\textrm{t}_1, I^\textrm{w}_2, I^\textrm{t}_2, \ldots, I^\textrm{w}_m])\textrm{,}
\end{align*}

\vspace{-0.25cm}
\noindent where the domain of the variables $C_i$ are the nodes $\mathbb{C}$ of the active component's roadmap $r = (\mathbb{C},\mathbb{E})$.
An interval variable $I^\textrm{w}_i$ models the time spent at configuration $C_i$ whereas $I^\textrm{t}_i$ denotes the traveling time between the configurations $C_i$ and $C_{i+1}$. For this purpose, the roadmap edges $\mathbb{E}$ denote the expected traveling time as edge weight. The roadmap thus yields information about
paths between configurations and their durations to be used in the CP. Note that the roadmaps may also include multiple nodes for the same configuration -- for example to consider the different collision geometries of the arm depending on the gripper state.
The set of all motion series -- one for each active component -- is named $\Sigma$.
Collisions between any two active components $a_i$ and $a_j$ are prevented by a constraint requiring that pairs of conflicting joint configurations $c_i$ and $c_j$, which are precomputed in a \emph{collision table}, must not be assumed simultaneously.

\subsection{Connection Variables}
\noindent The task model does not contain any information about the actual joint configurations and thus motions of the active components. Conversely, the motion model has no information on the OVCs, primitive actions, locations and resources. Yet, the two submodels are linked in two ways: First, for each active component, the \emph{location mapping} links the active component's roadmap nodes $\mathbb{C}$ with the locations. That is, for each location $l \in \mathbb{L}$, it provides a set of the joint configurations $c_l \subset \mathbb{C}$ that reach $l$. Second, the \emph{Connection Variables} link from the location variables of the task model to configuration variables of the motion model. Such a connection denotes that the configuration variable has to be chosen such that the respective active component reaches the location given in the location variable. There exists exactly one such connection per location variable. Configuration variables not referenced by Connection Variables are used for evasive movements to avoid collisions and deadlocks.


The key idea is that these connections are also CP variables. Therefore the name Connection Variables. They can be considered as meta variables, as they specify to which configuration variable to point to. Hence, formally, the domain of the Connection Variable $X_{\omega,j}$ for the location variable $L_j$ of OVC $\omega$ is the index of the configurations $[C_1, \ldots, C_m]$ of the motion series $\sigma$ of the active component $A$ of $\omega$. We refer to this mechanism as \emph{index-based}.
An assignment $X_{\omega,j} = i$ states that the $i$th configuration variable $C_i$ of the motion series $\sigma$ of $A$ has to reach to the $j$th location of $\omega$, formally $C_i \in \lambda(L_j)$. In this way, the Connection Variables establish the execution order for the OVCs assigned to an active component. 

Connection Variables of $\omega$ always have to be strictly monotonic, i.e. $X_{\omega,j} < X_{\omega,j+1}$, since the locations $[L_1, \ldots, L_l]$ of $\omega$ have to be visited in this order.
Yet, two OVCs for the same active component may be interleaved (e.g.,~$X_{\omega_1,1} = 3$, $X_{\omega_2,1} = 4$, $X_{\omega_2,2} = 5$, and $X_{\omega_1,2} = 6$) if there is no conflicting inter-OVC constraint or resource-constraint.
Two Connection Variables must never reference the same configuration variable.


\subsection{Examples for Task Modeling with OVCs}\label{sec:ovc-example}

In the following, we exemplary define two parts of the molding use-case. First, we model the electrical check which needs synchronized behavior of both arms. Second, we sketch a bi-manual pick of a workpiece container.


For the electrical check in the molding use-case, two actions have to be coordinated. This task is modeled by two OVCs: $\omega_\textrm{fixture}$ for holding the part into the fixture and $\omega_\textrm{push}$ to push the button for starting the check. $\omega_\textrm{push}$ has one location variable constrained to the button location (i.e.\ $L_1 = l_\textrm{Button}$). $\omega_\textrm{fixture}$ has three location variables constrained to the pick location of the object,
the fixture location, and the destination container. To synchronize the two OVCs, we constrain the push interval ($I_1$ of $\omega_\textrm{push}$) to be during the fixture interval ($I_2$ of $\omega_\textrm{fixture}$). 



To model the bi-manual pick up of a workpiece container in the molding use-case, we create two OVCs $\omega_\textrm{L}$ and $\omega_\textrm{R}$, each of length $2$. We constrain the first location variables $L_1$ to assume either of the locations $\{l_\textrm{Grasp1}, l_\textrm{Grasp2}$\}. Through an inter-OVC constraint, we ensure that the combination of selected grasp poses yields a valid combination for a stable grasp. We constrain the second location variables $L_2$ to take either of the locations $\{l_\textrm{Place1}, l_\textrm{Place2}$\} and an additional inter-OVC constraint ensures, that all $4$ pick and place locations take compatible values. Temporal constraints ensure that the first intervals $I_1$ end together and the second intervals $I_2$ start together. Note, that the arms have to be controlled by a dedicated controller for the actual carrying. We assume, that the controller either provides information about occupied space over time, such that the solver can schedule possible other components to not interfere, or stays within a given subset of the workspace.

\section{STAAMS Solver for OVCs}\label{sec:ovc-planning}
\noindent In the following, we explain the horizon estimation for the number of configurations in each motion series. Then, we describe the overall search strategy, detailing the choices for variable and value selection heuristics.

\subsection{Horizon estimation}
\noindent Initially, we do not know the optimal horizon $m$ for every active component, i.e. the number of configuration variables. Therefore, we integrate an iterative deepening approach directly in our model.
For each active component $a$, we create a constraint variable $H$ named \emph{horizon}. We prevent any movements after the $H$th configuration in the motion series of $a$ by constraining all configurations $C_{i > H}$ to $C_H$.
Small horizon values generally render the problem unsatisfiable, while large values bloat the search space unnecessarily and may cause superfluous motions. A lower bound for $H$ is the number of all location variables of the OVCs assigned to the corresponding active components. Therefore, the placement of the horizon variables in the search strategy is important, which is explained in the following Section. 

\begin{figure*}[!t]
\centering
\includegraphics[width=0.95\textwidth]{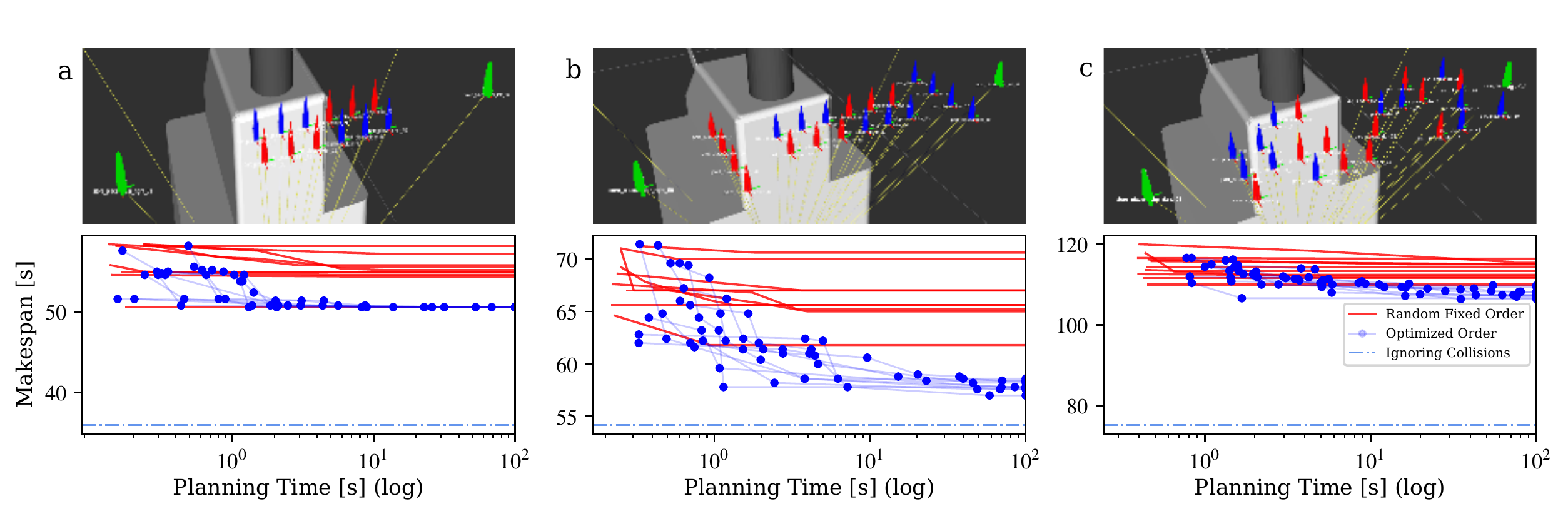}
\vspace{-0.4cm}
\caption{Sorting scenarios (a)-(c) and makespan-vs-planning-time plots. Red lines show the makespan over planning time for a random fixed order of execution (cf.~\cite{kimmel_scheduling_2016}). The blue lines depict the makespan, when we let the solver decide on the order. A lower bound for each problem -- obtained by ignoring collisions (relaxation of the problem) -- is plotted in light blue, 
Blue Objects are dropped into a container by the left arm at the left destination (green), and vice versa for the red objects. (a) $12$ objects with high conflict potential, (b) as (a) but with eight uncritical objects more to allow for efficient scheduling. (c) A randomly chosen instance with $24$ objects and much interaction}
\vspace{-0.6cm}
\label{fig:t-vs-q-a}
\end{figure*}

\subsection{Search strategy}
\noindent A constraint satisfaction solver computes one or more variable assignments that each satisfy all constraints. Such solvers usually interleave a backtracking search with constraint propagation. In the backtracking search, variables are selected according to a variable-ordering heuristic, and values for the variables are chosen based on a value-ordering heuristic. 

\textbf{Variable ordering.} The Connection Variables constitute a special case in our model. Due to their index-based mechanism, the constraint information from the motion model to the task model and vice-versa cannot be propagated until decisions on the involved Connection Variables have been made. The Connection Variables, again, require to first decide on the active component variables $A$ and the location variables $L_i$.
In our use-cases, searching (1.) on the location variables, (2.) on the active component variables, (3.) on the Connection Variables, (4.) on the  horizon variables, and then (5.) on the configurations variables of the motion series yielded reliably good solutions within a few seconds planning time. This order makes sure, that all constraints are added to the model before any time is spent on the actual motion scheduling. However, our approach allows the user to customize or further refine the search strategy. This includes the behavior within those five variable batches or the overall order, which is further discussed in Section~\ref{sec:evaluation}.
%
%
%
At this stage, only the the time interval variables of the resources, OVCs and motion series remain to be decided. As the time interval variables of the resources are connected to the OVCs, which in turn are linked with the motion series by the Connection Variables, the solver has to decide about the time-scaling of the motion series. More precisely, the solver has to decide about the waiting times $I^\textrm{w}_1, \ldots, I^\mathrm{w}_H$ of each motion series. The time-scaling allows to prevent collisions, resolve resource conflicts, and satisfy any inter-OVC constraint (e.g. synchronization or ordering). Also, no superfluous waiting times should be added to optimize the makespan. By solving this time-scaling problem as final step, we obtain the timed motion plans for all active components.

 
\textbf{Value selection.} For each selected variable, the solver has to assign a value from the variable's domain. In case of the Connection Variables (4) and horizon variables (3), we use a minimum value heuristic to foster short motion series. For (1), (2), and (5), we use a random value selection heuristic as there is no clear preference for these variables. In case of a good value selection, the remaining search process involves only few backtracks. We employ a Luby restart strategy (cf.~\cite{luby_optimal_1993}) to avoid long-lasting searches in the time-scaling step (i.e.~in the 6th~step). This is especially useful when poor value selections have been made in steps (1) to (5).

%
\begin{figure}[tpb]
\centering
\includegraphics[width=0.9\linewidth]{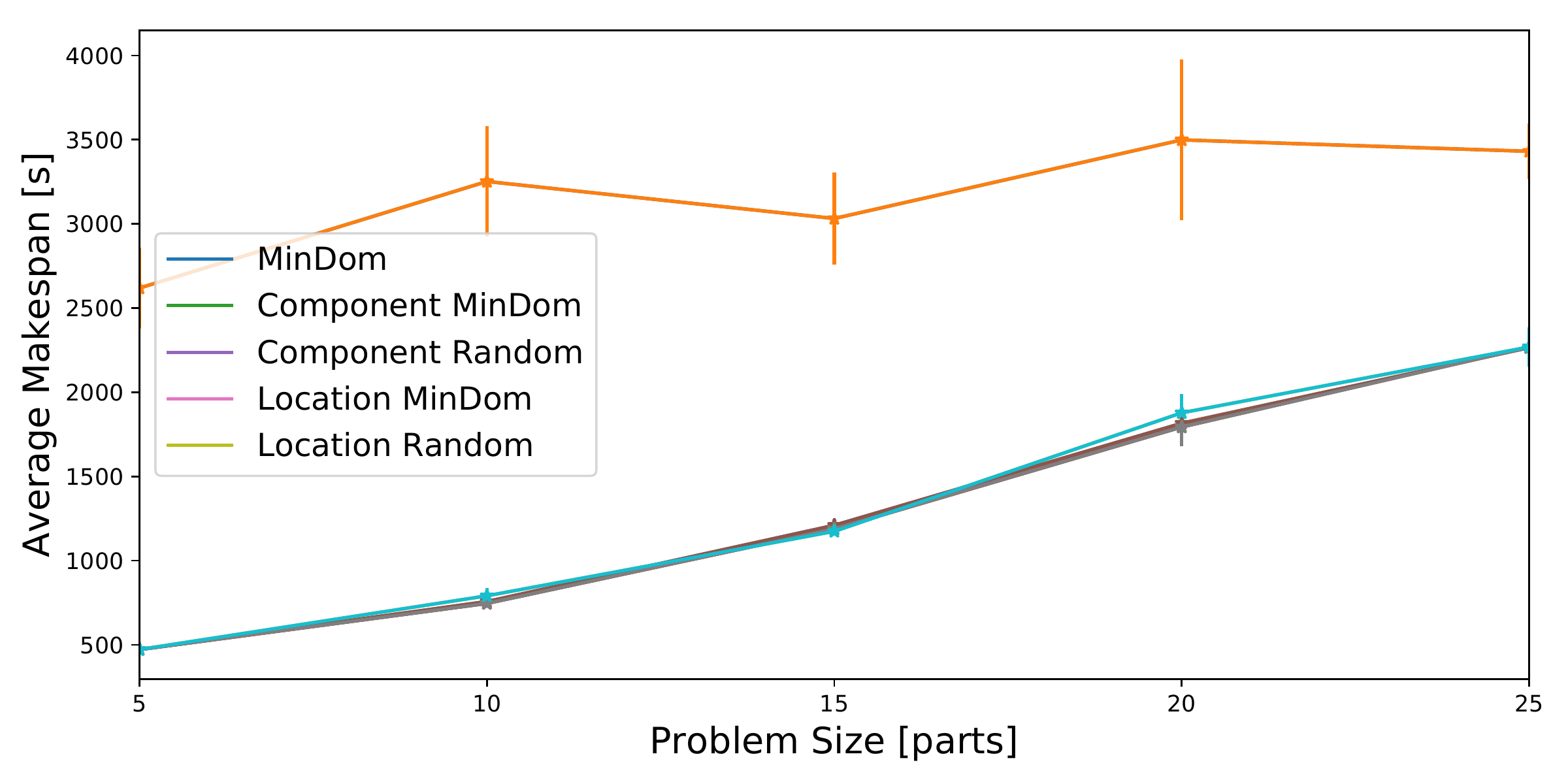}
\vspace{-0.4cm}
\caption{Aggregated plot over $125$ experiments showing the influence of the search strategy on the solution quality after $100~s$ for different problem sizes.}
\vspace{-0.6cm}
\label{fig:strategy_statistics}
\end{figure}
%


\section{Evaluation} \label{sec:evaluation}
\noindent We implemented our STAAMS model and solver in Python using the Google Operation Research Tools \cite{google_2018} and experimented with a KaWaDa Nextage dual-arm robot in a Gazebo simulation environment \cite{koenig_design_2004} on a HP zBook laptop. We implemented the first and second use-case given in Sec.~\ref{subsec:use-case analysis:use-case}. Here, we focus on the sorting use-case, which resembles the experiment by Kimmel at al.\ for their dual-arm coordination algorithm \cite{kimmel_scheduling_2016}, and compare the results. Afterwards, we show how our solver scales on instances of this use-case for up to $200$ objects. The modularity of our STAAMS model (cf.~Fig.~\ref{fig:Overview to our model}) enables the re-use of tasks expressed as \emph{Ordered Visiting Constraints}. To show this, we deployed an example task (taking all objects from a table) on two robotic platforms by reusing the task model.


\textbf{Comparison with pure time-scaling.} We modeled three instances of the sorting use case with increasing number of objects from \numrange{12}{24} and varying degree of conflict between the two arms (see Fig.~\ref{fig:t-vs-q-a}). Then, we compared our approach against the theoretical lower bound obtained by ignoring collisions between the manipulators as well as against the method by Kimmel et al., which time-scales the trajectories of both manipulators to prevent collisions. We mimic their solver by using a randomized but fixed order of collecting the objects and leave only the scheduling to our solver. The results are visualized in Fig.~\ref{fig:t-vs-q-a}. The diagrams show plots of the makespan (as quality measure) over the time spent to solve the instance (stopped after \SI{100}{\second}) for ten different fixed order runs (red) and ten runs with order optimization (dark-blue). Our solver produces the first solutions sometimes as fast as in \SI{0.1}{\second} and usually converges within \SI{3}{\second} on the instances shown. By optimizing the order, our solver consistently outperforms the fixed order runs -- or reaches the same performance in the rare case that by chance a very good order is selected. Since both approaches utilize some random decisions, the plotted outcomes visualize a distribution. With this in mind, it becomes very clear that our STAAMS solver provides much more consistent and higher-quality results. In scenario (b), it gets very close to the theoretical lower bound (light blue). Interestingly, it takes only \SI{7}{\second} more to handle eight additional objects in (b) compared to (a).


\begin{figure}[!t]
\centering
\input{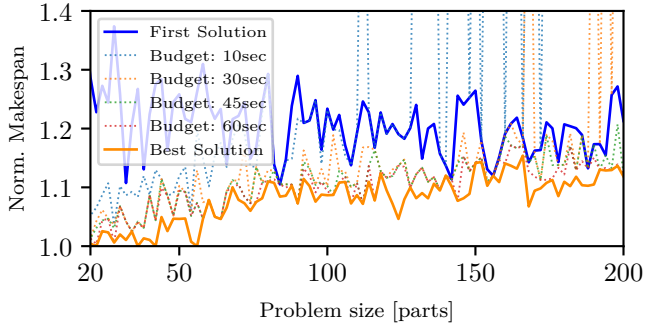}
\vspace{-1.1cm}
\caption{In this diagram, the solution quality (makespan) divided by the lower bound (which ignores collisions between the manipulators) is shown for different problem sizes and different stages in the search. The vertical lines in the right half of the figure indicate cases in which no solution was found within the budgets of $10$ or rather $30\,\mathrm{sec}$.}
\vspace{-0.75cm}
\label{fig:quality-vs-problem-size}
\end{figure}

\textbf{Scalability.} To evaluate the scaling properties of our approach, we ran a series of $80$ experiments similar to scenario (b) with a time limit of $\SI{180}{\second}$. Starting from the twenty parts depicted in \figref{fig:t-vs-q-a}b, we added for each experiment two extra parts to the scene -- one for each arm -- up to $200$ parts in total. In \figref{fig:quality-vs-problem-size}, the normalized makespan, i.e. the makespan divided by the theoretical lower bound, is plotted over the problem size for the first solutions, the best solutions, and computing time budgets from $10$ to $\SI{60}{\second}$. The solution quality for the first solution ranges approximately from $1.1$ to $1.37$ normalized makespan
(solutions with a normalized makespan of approx. $2$ can be constructed),
which rapidly improves with the following solutions to finally settle around $1.1$ normalized makespan. 


Our solver computes high-quality solutions even for large problem instances in a few seconds or tens of seconds. Please note that the high scalability compared to ITAMP planners stems from two facts: First, STAAMS solving does not require to decide about the actions to be executed but rather to complete and optimize a given abstract plan (here modeled by OVCs) only. Second, in our motion model we limit the motions to stick to predefined roadmaps.

\begin{figure}[tpb]
\centering
\includegraphics[width=0.99\linewidth]{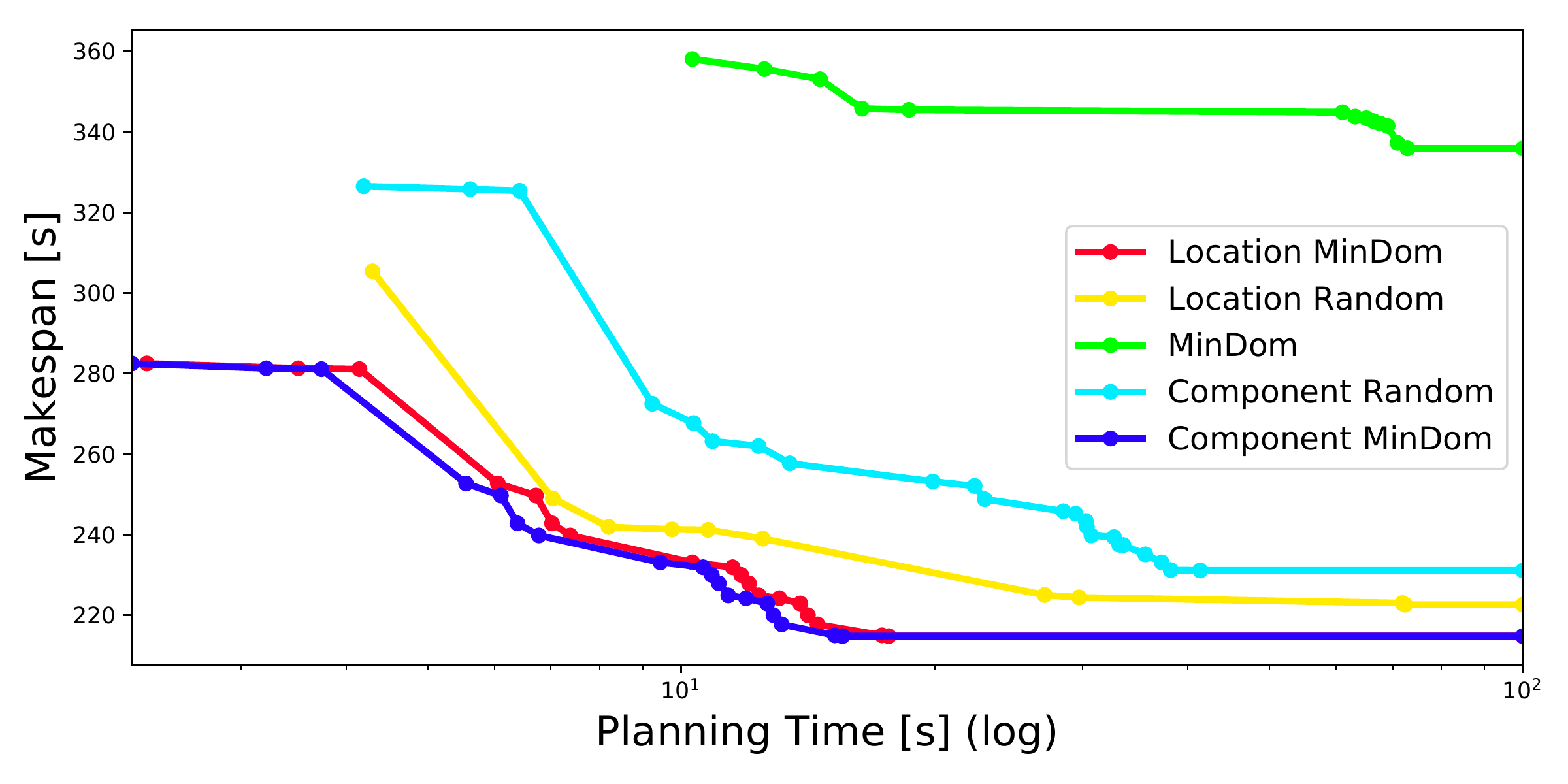}
\vspace{-0.8cm}
\caption{Makespan vs. planning time for different search strategies on a $25$ OVC task.}
\vspace{-0.3cm}
\label{fig:strategy_makespan_time}
\end{figure}

\textbf{Custom search strategies.} In Fig.~\ref{fig:strategy_statistics}, we compared four strategies, which are compliant with the rules explained in Sec.~\ref{sec:ovc-planning}, with a general-purpose baseline strategy. For the baseline strategy, variables were selected using a minimum domain heuristic, while values were selected randomly. We ran a total of 125 experiments for $\SI{100}{\second}$ each. With each strategy, we solved five differently sized problems ($5-25$ OVCs). Each of these experiments were executed five times with varied random seed for the solver. Fig.~\ref{fig:strategy_makespan_time} shows how the five strategies perform over time on a problem with $25$ OVCs -- it can be seen that the strategy makes a significant difference for the convergence speed.
With these experiments we show, that although solutions can be found with the baseline strategy, the specific search strategies are necessary to achieve good solutions in acceptable solving time. In our use-cases, all custom strategies delivered comparable performance after $\SI{20}-\SI{30}{\second}$ of search (see Fig.~\ref{fig:strategy_makespan_time}). However, the differences for short solving times clearly show the benefit of employing a suitable search strategy. 

\begin{figure}[tpb]
\centering
\includegraphics[width=0.8\linewidth]{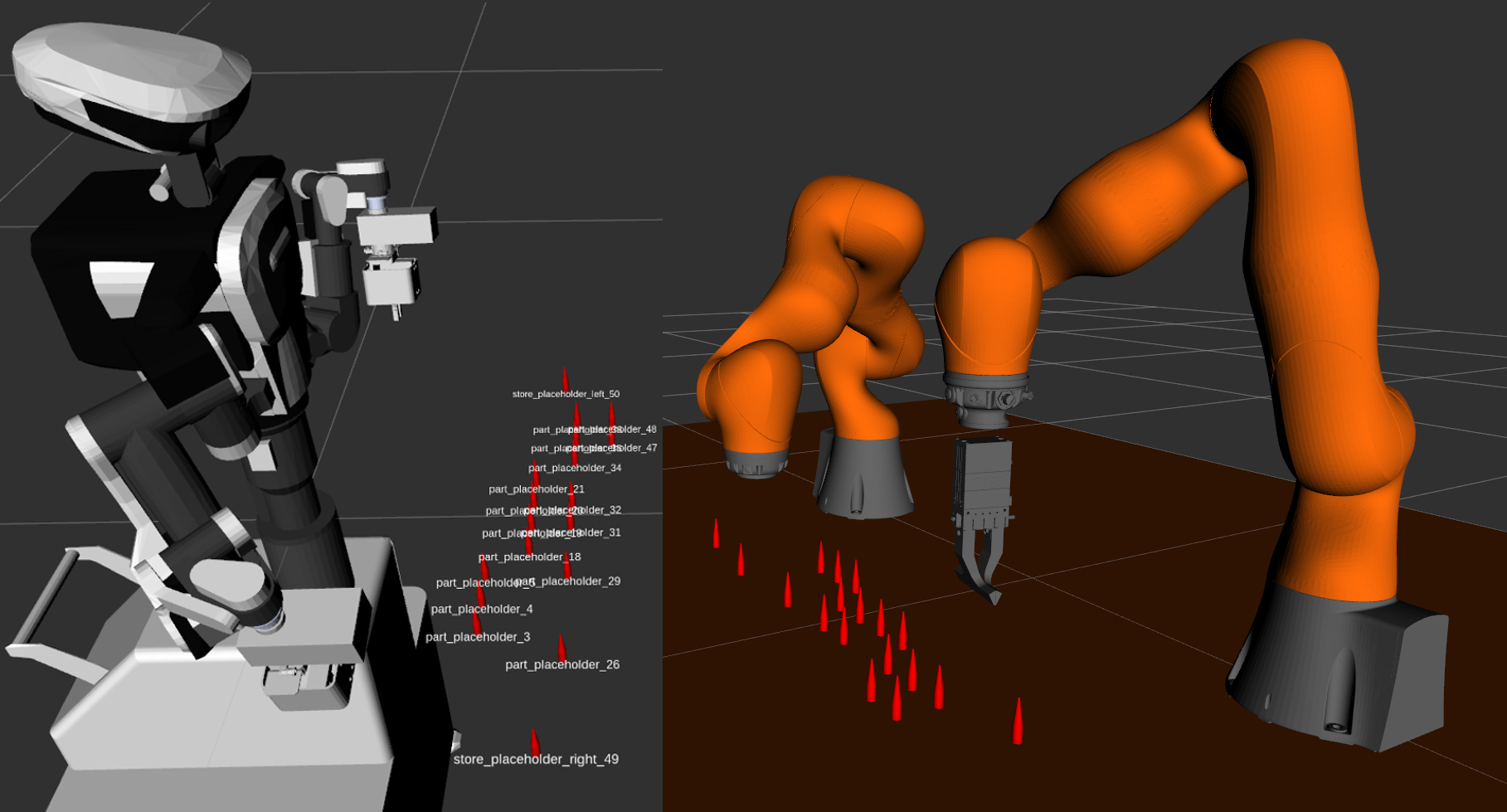}
\vspace{-0.2cm}
\caption{A task -- cleaning up the table -- deployed on a KaWaDa Nextage robot (left) and a pair of KUKA LBR iiwa robots (right).}
\vspace{-0.78cm}
\label{fig:portability}
\end{figure}

\textbf{Portability.} Flexibility and portability of our modeling language are validated through several experiments on different simulated robot platforms (see Fig.~\ref{fig:portability}). The task models, i.e. the sets of OVCs, that have been used to perform the pallet emptying on the KaWaDa Nextage and KUKA LBR iiwa platforms
are identical. The differences are:\footnote{The code and the setup details are available at \texttt{https://github.com/boschresearch/STAAMS-SOLVER}}
The robot model (Moveit! \cite{moveit_2018} robot configuration to access motion planning, kinematic calculations, and collision checking); a seed robot configuration (as required by the Inverse kinematics (ik) solvers); a "tuck" robot configuration (in which the arms do not obstruct each others workspaces); the names of kinematic chains, end-effectors, and the base frame; the static scene collision layout (represented in meshes or primitive shapes; and the locations of the workpieces. With this information and scripts in place, our system automatically creates the roadmaps, collision tables, and name-location-configuration mappings. 

\section{Conclusion and Future Work}\label{sec:conclusion}
\noindent In this paper, we proposed a flexible model and solver for simultaneous task allocation and motion scheduling (STAAMS) based on constraint programming (CP) and constraint optimization for industrial manipulation and assembly tasks for dual-arm robots. The core modeling concepts are Ordered Visiting Constraints and time-scalable motion series, which are linked by meta CP variables named Connection Variables. In our evaluation, we showed that our STAAMS solver quickly completes and optimizes a given problem model instance -- i.e. an abstract task specifications given as collection of OVCs for a robot motion model -- and delivers high-quality,  executable motion plans. We demonstrated the scalability of our approach on large problem instances with up to $200$ actions, which were solved in less than $\SI{180}{\second}$. We also showed, that the OVC concept allows to transfer a given task model to another robot and/or workspace by exchanging the relevant motion submodels only. 
We assume that our task-centric robot programming approach is suited not only for textual specification but also for multi-modal input variants. Therefore, we want to explore robot programming by natural language and demonstrations.

To broaden the applicability of this approach, we plan to include more task primitives (additionally to reaching goals) like trajectories for welding. We will also investigate the extension of our approach to include action models with safe approximations, when the actual space occupancy and duration are not known, e.g., when employing force-position control.

\section{Acknoledgement}

\noindent The authors thank the project Robotics for Industry 4.0 (reg. no. CZ.02.1.01/0.0/0.0/15 003/0000470) for providing the KUKA robot simulation setup used in the Evaluation section.






\bibliographystyle{plain}
\bibliography{ms}

\end{document}